\documentclass[a4paper]{article}

\usepackage{INTERSPEECH2019}
\usepackage{amsmath,amsfonts,amssymb,amsthm}
\usepackage{times}
\usepackage{latexsym}
\usepackage{hyperref}
\usepackage{url}
\usepackage{booktabs}
\usepackage{times}
\usepackage{helvet}
\usepackage{courier}
\usepackage{amsmath, amssymb}
\usepackage{bm}
\usepackage{float}
\usepackage{graphicx}
\usepackage{caption}
\usepackage{subfig}
\usepackage{multirow}
\usepackage{algorithm2e}
\usepackage{enumitem}
\usepackage{fancyvrb}
\usepackage{relsize}
\usepackage{tipa}
\usepackage{tikz}
\usepackage{pgfplots}
\title{Multilingual Speech Recognition with Corpus Relatedness Sampling}

\name{Xinjian Li, Siddharth Dalmia, Alan W. Black, Florian Metze}
\address{Language Technologies Institute, Carnegie Mellon University; Pittsburgh, PA; U.S.A.
\email{\{xinjianl|sdalmia|awb|fmetze\}@cs.cmu.edu}}
\begin{document}
\maketitle
\begin{abstract}
Multilingual acoustic models have been successfully applied to low-resource speech recognition. Most existing works have combined many small corpora together, and pretrained a multilingual model by sampling from each corpus uniformly. The model is eventually fine-tuned on each target corpus. This approach, however, fails to exploit the relatedness and similarity among corpora in the training set. For example, the target corpus might benefit more from a corpus in the same domain or a corpus from a close language. In this work, we propose a simple but useful sampling strategy to take advantage of this relatedness. We first compute the corpus-level embeddings and estimate the similarity between each corpus. Next we start training the multilingual model with uniform-sampling from each corpus at first, then we gradually increase the probability to sample from related corpora based on its similarity with the target corpus. Finally the model would be fine-tuned automatically on the target corpus. Our sampling strategy outperforms the baseline multilingual model on 16 low-resource tasks. Additionally, we demonstrate that our corpus embeddings capture the language and domain information of each corpus.
\end{abstract}

\section{Introduction}
\label{sec:intro}

In recent years, Deep Neural Networks (DNNs) have been successfully applied to Automatic Speech Recognition (ASR) for many well-resourced languages including Mandarin and English \cite{amodei2016deep, xiong2016achieving}. However, only a small portion of languages have clean speech labeled corpus. As a result, there is an increasing interest in building speech recognition systems for low-resource languages. To address this issue, researchers have successfully exploited multilingual speech recognition models by taking advantage of labeled corpora in other languages \cite{huang2013cross, heigold2013multilingual}. Multilingual speech recognition enables acoustic models to share parameters across multiple languages, therefore low-resource acoustic models can benefit from rich resources.

While low-resource multilingual works have proposed various acoustic models, those works tend to combine several low-resource corpora together without paying attention to the variety of corpora themselves. One common training approach here is to first pretrain a multilingual model by combining all training corpora, then the pretrained model is fine-tuned on the target corpus \cite{dalmia2018sequence}. During the training process, each corpus in the training set is treated equally and sampled uniformly. We argue, however, this approach does not take account of the characteristics of each corpus, therefore it fails to take advantage of the relations between them. For example, a conversation corpus might be more beneficial to another conversation corpus rather than an audio book corpus.

In this work, we propose an effective sampling strategy (Corpus Relatedness Sampling) to take advantage of relations among corpora. Firstly, we introduce the corpus-level embedding which can be used to compute the similarity between corpora. The embedding can be estimated by being jointly trained with the acoustic model. Next, we compute the similarity between each corpus and the target corpus, the similarity is then used to optimize the model with respect to the target corpus. During the training process, we start by uniformly sampling from each corpus, then the sampling distribution is gradually updated so that more related corpora would be sampled more frequently. Eventually, only the target corpus would be sampled from the training set as the target corpus is the most related corpus to itself. While our approach differs from the pretrained model and the fine-tuned model, we can prove that those models are special cases of our sampling strategy.

To evaluate our sampling strategy, we compare it with the pretrained model and fine-tuned model on 16 different corpora. The results show that our approach outperforms those baselines on all corpora: it achieves 1.6\% lower phone error rate on average. Additionally, we demonstrate that our corpus-level embeddings are able to capture the characteristics of each corpus, especially the language and domain information. The main contributions of this paper are as follows:
\begin{enumerate}
    \item We propose a corpus-level embedding which can capture the language and domain information of each corpus.
    \item We introduce the Corpus Relatedness Sampling strategy to train multilingual models. It outperforms the pretrained model and fine-tuned model on all of our test corpora.
\end{enumerate}

\section{Related Work}


Multilingual speech recognition has explored various models to share parameters across languages in different ways. For example, parameters can be shared by using posterior features from other languages \cite{stolcke2006cross}, applying the same GMM components across different HMM states \cite{burget2010multilingual}, training shared hidden layers in DNNs \cite{huang2013cross, heigold2013multilingual} or LSTM \cite{dalmia2018sequence}, using language independent bottleneck features \cite{vesely2012language, dalmia2018domain}. Some models only share their hidden layers, but use separate output layers to predict their phones \cite{huang2013cross, heigold2013multilingual}. Other models have only one shared output layer to predict the universal phone set shared by all languages \cite{schultz2001language, tong2017investigation, vu2013multilingual}. While those works proposed the multilingual models in different ways, few of them have explicitly exploited the relatedness across various languages and corpora. In contrast, our work computes the relatedness between different corpora using the embedding representations and exploits them efficiently.

The embedding representations have been heavily used in multiple fields. In particular, embeddings of multiple granularities have been explored in many NLP tasks. To name a few, character embedding \cite{kim2016character}, subword embedding \cite{bojanowski2017enriching}, sentence embedding \cite{kiros2015skip} and document embedding \cite{le2014distributed}. However, there are few works exploring the corpus level embeddings. The main reason is that the number of corpora involved in most experiments is usually limited and it is not useful to compute corpus embeddings. The only exception is the multitask learning where many tasks and corpora are combined together. For instance, the language level (corpus level) embedding can be generated along with the model in machine translation \cite{johnson2017google} and speech recognition \cite{li2018multi}. However, those embeddings are only used as an auxiliary feature to the model, few works continue to exploit those embeddings themselves.

Another important aspect of our work is that we focused on the sampling strategy for speech recognition. While most of the previous speech works mainly emphasized the acoustic modeling side, there are also some attempts focusing on the sampling strategies. For instance, curriculum learning would train the acoustic model by starting from easy training samples and increasingly adapt it to more difficult samples \cite{amodei2016deep, braun2017curriculum}. Active learning is an approach trying to minimize human costs to collect transcribed speech data \cite{riccardi2005active}. Furthermore, sampling strategies can also be helpful to speed up the training process \cite{cui2015multilingual}. However, the goals of most strategies are to improve the acoustic model by modifying the sampling distribution within a single speech corpus for a single language. On the contrary, our approach aims to optimize the multilingual acoustic model by modifying distributions across all the training corpora.

\section{Approach}
In this section, we describe our approach to compute the corpus embedding and our Corpus Relatedness Sampling strategy.

\begin{table*}[t]
\begin{center}
    \caption{The collection of training corpora used in the experiment. Both the baseline model and the proposed model are trained and tested with 16 corpora across 10 languages. We assign a corpus id to each corpus after its corpus name so that we can distinguish the corpora sharing the same language.} \label{tab:corpus}
    \begin{tabular}{c c c c | c c c c} 
    \toprule
    {\bf Language} & {\bf Corpus Name} & {\bf Domain} & {\bf Utterance} & {\bf Language } & {\bf Corpus Name }  & {\bf Domain} & {\bf Utterance}\\
    \midrule
    English & TED (ted) \cite{rousseau2012ted} & broadcast & 100,000  & Mandarin & Hkust (hk) \cite{liu2006hkust} & telephone & 100,000 \\
    English & Switchboard (swbd)\cite{godfrey1992switchboard} & telephone & 100,000 & Mandarin & SLR18 (s18) \cite{THCHS30_2015} & read & 13,388 \\
    English & Librispeech (libri) \cite{panayotov2015librispeech} & read & 100,000  & Mandarin & LDC98S73 (hub) & broadcast & 35,999 \\
    English & Fisher (fisher) \cite{cieri2004fisher} & telephone & 100,000 & Mandarin & SLR47 (s47) \cite{primewords_201801} & read & 50,384 \\
    \midrule
    Amharic & LDC2014S06 (babel) & telephone & 41,403 & Bengali & LDC2016S08 (babel) & telephone & 60,663 \\
    Dutch & voxforge (vox) & read  & 8,492 & German & voxforge (vox) & read & 41,146 \\
    Spanish & LDC98S74 (hub) & broadcast & 31,615 & Swahili & LDC2017S05(babel) & telephone & 44,502 \\
    Turkish & LDC2012S06 (hub) \cite{saraclar2012turkish} & broadcast & 97,427 & Zulu & LDC2017S19 (babel) & telephone & 60,835 \\
    \bottomrule
    \end{tabular}
\end{center}
\end{table*}

\subsection{Corpus Embedding}
Suppose that $\mathcal{C}_t$ is the target low-resource corpus, we are interested in optimizing the acoustic model with a much larger training corpora set $\mathcal{S} = \{ \mathcal{C}_1, \mathcal{C}_2 ... \mathcal{C}_n \}$ where $n$ is the number of corpora and $\mathcal{C}_t \in \mathcal{S}$. Each corpus $\mathcal{C}_i$ is a collection of $(\mathbf{x}, \mathbf{y})$ pairs where $\mathbf{x}$ is the input features and $\mathbf{y}$ is its target.

\begin{figure}[h]
  \centering
  \includegraphics[width=0.7\linewidth]{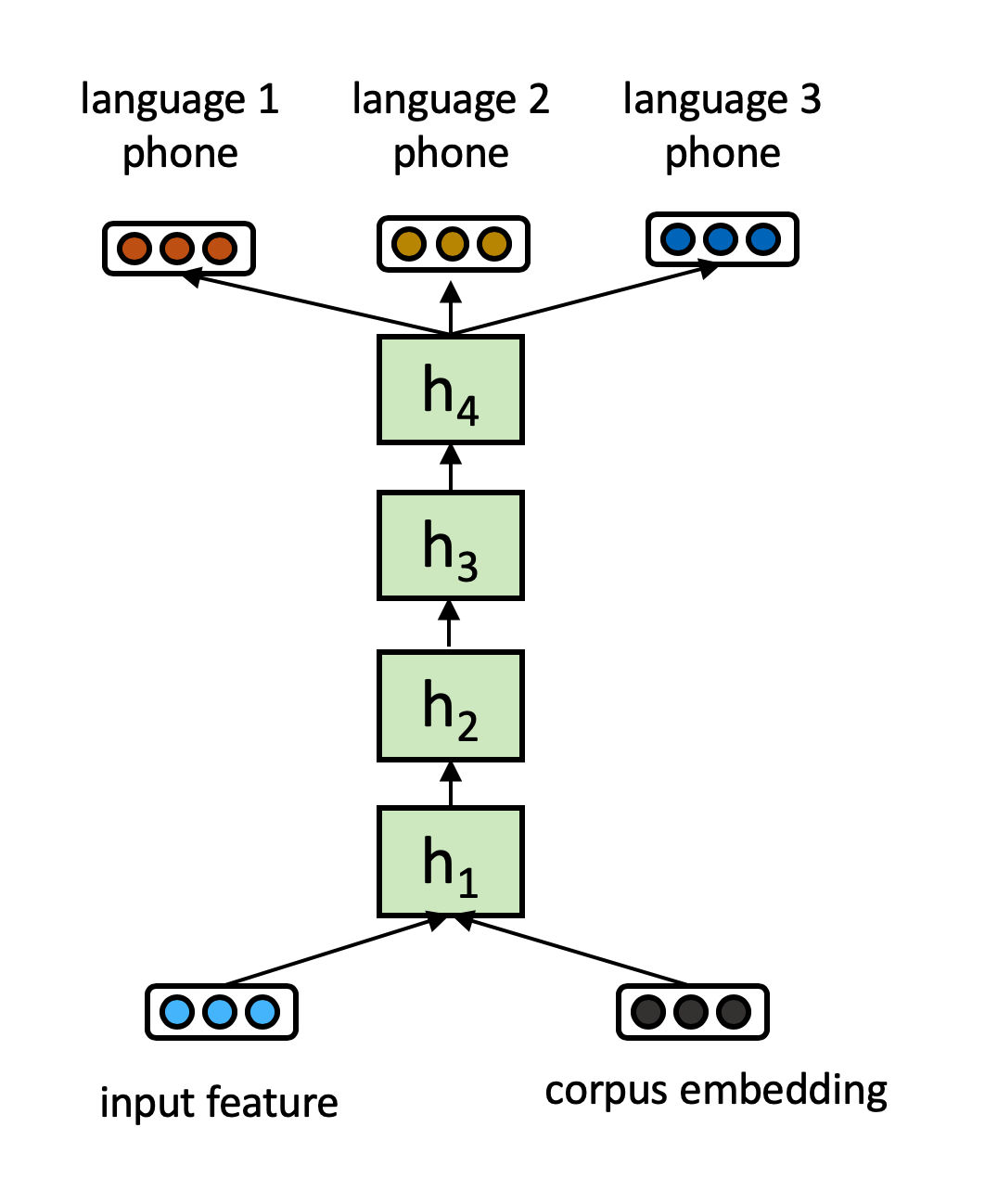}
  \caption{The acoustic model to optimize corpus embeddings.}
  \label{fig:arch}
\end{figure}

Our purpose here is to compute the embedding $\mathbf{e}_i$ for each corpus $\mathcal{C}_i$ where $\mathbf{e}_i$ is expected to encode information about its corpus $\mathcal{C}_i$. Those embeddings can be jointly trained with the standard multilingual model \cite{dalmia2018sequence}. First, the embedding matrix $E$ for all corpora is initialized, the $i$-th row of $E$ is corresponding to the embedding $\mathbf{e}_i$ of the corpus $\mathcal{C}_i$. Next, during the training phase, $\mathbf{e}_i$ can be used to bias the input feature $\mathbf{x}$ as follows.

\begin{align}
                \mathbf{h} = \mathrm{Encoder}(\mathbf{x}+\mathbf{e}_i; W, E)
\end{align}

where $(\mathbf{x}, \mathbf{y}) \in \mathcal{C}_i$ is an utterance sampled randomly from $\mathcal{S}$, $\mathbf{h}$ is its hidden features, $W$ is the parameter of the acoustic model and Encoder is the stacked bidirectional LSTM as shown in Figure.\ref{fig:arch}. Next, we apply the language specific softmax to compute logits $\mathbf{l}$ and optimize them with the CTC objective \cite{graves2006connectionist}. The embedding matrix $E$ can be optimized together with the model during the training process.

\subsection{Corpus Relatedness Sampling}
With the embedding $\mathbf{e}_i$ of each corpus $\mathcal{C}_i$, we can compute the similarity score between any two corpora using the cosine similarity.

\begin{align}
    score(\mathcal{C}_i, \mathcal{C}_j) = \frac{\mathbf{e}_i \cdot \mathbf{e}_j}{ | \mathbf{e}_i | |\mathbf{e}_j|}
\end{align}

As the similarity reflects the relatedness between corpora in the training set, we would like to sample the training set based on this similarity: those corpora which have a higher similarity with the target corpus $\mathcal{C}_t$ should be sampled more frequently. Therefore, we assume those similarity scores to be the sampling logits and they should be normalized with softmax.

\begin{align}
    Pr(\mathcal{C}_i) = \frac{\exp\big(T \cdot score(\mathcal{C}_i, \mathcal{C}_t)\big)}{\sum_j{\exp(T \cdot score(\mathcal{C}_j, \mathcal{C}_t))}}
\end{align}

where $Pr(\mathcal{C}_i)$ is the probability to sample $\mathcal{C}_i$ from $\mathcal{S}$, and $T$ is the temperature to normalize the distribution during the training phase. We argue that different temperatures could create different training conditions. The model with a lower temperature tends to sample each corpus equally like uniform sampling. In contrast, a higher temperature means that the sampling distribution should be biased toward the target corpus like the fine-tuning. 

Next, we prove that both the pretrained model and the fine-tuned model can be realized with specific temperatures. In the case of the pretrained model, each corpus should be sampled equally. This can be implemented by setting $T$ to be $0$.
\begin{align}
    Pr(\mathcal{C}_i) = \lim_{T \to 0} \frac{\exp\big(T \cdot score(\mathcal{C}_i, \mathcal{C}_t)\big)}{\sum_j{\exp(T \cdot score(\mathcal{C}_j, \mathcal{C}_t))}} = \frac{1}{n}
\end{align}

On the other hand, the fine-tuned model should only consider samples from the target corpus $\mathcal{C}_t$ , while ignoring all other corpora. We argue that this condition can be approximated by setting $T$ to a very large number. As $score(\mathcal{C}_t, \mathcal{C}_t) = 1.0$ and $score(\mathcal{C}_i, \mathcal{C}_t) < 1.0$ if $i \neq t$, we can prove the statement as follows:

\begin{equation}
\begin{split}
Pr(\mathcal{C}_i) & = \lim_{T \to \infty}  \frac{\exp\big(T \cdot score(\mathcal{C}_i, \mathcal{C}_t)\big)}{\sum_j{\exp(T \cdot score(\mathcal{C}_j, \mathcal{C}_t))}} \\ 
& = \begin{cases}
1.0 & if\> \mathcal{C}_i = \mathcal{C}_t \\
0.0 & if\> \mathcal{C}_i \neq \mathcal{C}_t
\end{cases}
\end{split}
\end{equation}

While both the pretrained model and the fine-tuned model are special cases of our approach, we note that our approach is more flexible to sample from related corpora by interpolating between those two extreme temperatures. In practice, we would like to start with a low temperature to sample broadly in the early training phase. Then we gradually increase the temperature so that it can focus more on the related corpora. Eventually, the temperature would be high enough so that the model is automatically fine-tuned on the target corpus. Specifically, in our experiment, we start training with a very low temperature $T_0$, and increase its value every epoch $k$ as follows.

\begin{align}
T_{k+1} = a T_{k}
\end{align}
where $T_k$ is the temperature of epoch $k$ and $a$ is a hyperparameter to control the growth rate of the temperature.

\section{Experiments}
To demonstrate that our sampling approach could improve the multilingual model, we conduct experiments on 16 corpora to compare our approach with the pretrained model and fine-tuned model. 

\subsection{Datasets}
We first describe our corpus collection. Table.\ref{tab:corpus} lists all corpora we used in the experiments. There are 16 corpora from 10 languages. To increase the variety of corpus, we selected 4 English corpora and 4 Mandarin corpora in addition to the low resource language corpora. As the target of this experiment is low resource speech recognition, we only randomly select 100,000 utterances even if there are more in each corpus. All corpora are available in LDC, voxforge, openSLR or other public websites. Each corpus is manually assigned one domain based on its speech style. Specifically, the domain candidates are \textit{telephone}, \textit{read} and \textit{broadcast}.

\subsection{Experiment Settings}
We use EESEN \cite{miao2015eesen} for the acoustic modeling and epitran \cite{mortensen2018epitran} as the g2p tool in this work. Every utterance in the corpora is firstly re-sampled into 8000Hz, and then we extract 40 dimension MFCCs features from each audio. We use a recent multilingual CTC model as our acoustic architecture \cite{dalmia2018sequence}: The architecture is a 6 layer bidirectional LSTM model with 320 cells in each layer. We use this architecture for both the baseline models and the proposed model.

Our baseline model is the fine-tuned model:  we first pretrained a model by uniformly sampling from all corpora. After the loss converges, we fine-tune the model on each of our target corpora. To compare it with our sampling approach, we first train an acoustic model to compute the embeddings of all corpora, then the embeddings are used to estimate the similarity as described in the previous section. The initial temperature $T_0$ is set to 0.01, and the growth rate is $1.5$. We evaluated all models using the phone error rate (PER) instead of the word error rate (WER). The reason is that we mainly focus on the acoustic model in this experiment. Additionally, some corpora (e.g.: Dutch voxforge) in this experiment have very few amounts of texts, therefore it is difficult to create a reasonable language model without augmenting texts using other corpora, which is beyond the scope of this work.

\subsection{Results}

\begin{table}[h]
\begin{center}
\caption{Phone error rate (\%PER) of the pretrained model, the baseline model (fine-tuned model) and our CRS(Corpus Relatedness Sampling) approach on all 16 corpora}
\begin{tabular}{l c c c}
\toprule 
{\bf Corpus} & {\bf Pretrain PER}& {\bf Base PER} & {\bf CRS PER} \\
\midrule
English (ted) & 19.2 & 11.6 & 10.3 \\
English (swbd) &  24.9 & 15.3 & 14.1 \\
English (libri) & 12.1 & 6.5 & 5.4 \\
English (fisher) & 34.7 & 23.5 & 22.5 \\
Mandarin (hk) & 32.6 & 16.1 & 14.5 \\
Mandarin (s18) & 8.5 & 6.6 & 5.6 \\
Mandarin (hub) & 10.2 & 5.1 & 4.4 \\
Mandarin (s47) & 13.0 & 10.9 & 9.2 \\
\midrule
Amharic (babel) & 45.4 & 40.7 & 36.9 \\
Bengali (babel) &  47.0 & 41.9 & 40.0 \\
Dutch (vox) & 27.3 & 21.6 & 18.2 \\
German (vox) & 23.1 & 12.9 & 10.3 \\
Swahili (babel) & 17.7 & 16.1 & 14.5 \\
Spanish (hub) & 14.0 & 8.4 & 7.5 \\
Turkish (hub) & 49.2 & 45.6 & 44.9 \\
Zulu (babel) & 47.8 & 38.5 & 37.9 \\
\midrule
Average & 26.7 & 20.1 & \bf{18.5} \\
\bottomrule
\end{tabular}
\label{table:result}
\end{center}
\end{table}

Table.\ref{table:result} shows the results of our evaluation. We compare our approach with the baseline using all corpora. The left-most column of Table.\ref{table:result} shows the corpus we used for each experiment, the remaining columns are corresponding to the phone error rate of the pretrained model, the fine-tuned model and our proposed model. First, we can easily confirm that the fine-tuned model outperforms the pretrained model on all corpora. For instance, the fine-tuned model outperforms the pretrained model by 4.7\% on the Amharic corpus. The result is reasonable as the pretrained model is optimized with the entire training set, while the fine-tuned model is further adapted to the target corpus. Next, the table suggests our Corpus Relatedness Sampling approach achieves better results than the fine-tuned model on all test corpora. For instance, the phone error rate is improved from 40.7\% to 36.9\% on Amharic and is improved from 41.9\% to 40.0\% on Bengali. On average, our approach outperforms the fine-tuned model by 1.6\% phone error rate. The results demonstrate that our sampling approach is more effective at optimizing the acoustic model on the target corpus. We also train baseline models by appending corpus embeddings to input features, but the proposed model outperforms those baselines similarly.

One interesting trend we observed in the table is that the improvements differ across the target corpora. For instance, the improvement on the Dutch corpus is 3.4\%, on the other hand, its improvement of 0.6\% is relatively smaller on the Zulu dataset. We believe the difference in improvements can be explained by the size of each corpus. The size of Dutch corpus is very small as shown in Table.\ref{tab:corpus}, therefore the fine-tuned model is prone to overfit to the dataset very quickly. In contrast, it is less likely for a larger corpus to overfit. Compared with the fine-tuned model, our approach optimizes the model by gradually changing the temperature without quick overfitting. This mechanism could be interpreted as a built-in regularization. As a result, our model can achieve much better performance in small corpora by preventing the overfitting effect.

\begin{table}[h]
\begin{center}
\caption{The similarity between training corpora. For each target corpus, we show its most related corpus and second related corpus.}
\begin{tabular}{l | l | l}
\toprule 
{\bf Target} & {\bf 1st Related Corpus} & {\bf 2nd Related Corpus}\\
\midrule
English (ted) & English (libri) & Turkish (hub) \\
English (swbd) & English (fisher) & Mandarin (hk) \\
English (libri) & English (ted)&  German (vox) \\
English (fisher) & English (swbd) &Mandarin (hk) \\
Mandarin (hk) & Bengali (babel) &English (swbd) \\
Mandarin (s18) & Mandarin (s47)& Mandarin (hub) \\
Mandarin (s47) & Mandarin (s18)& Dutch (vox) \\
Mandarin (hub) & Spanish (hub)& Tkish (hub) \\
\midrule
Amharic (babel) & Bengali (babel) & Swahili (babel) \\
Bengali (babel) & Mandarin (hk)& Amharic (babel) \\
Dutch (vox) & Mandarin (s47) & Mandarin(s18) \\
German (vox) & English (libri) & Mandarin (s47) \\
Swahili (babel) & Zulu (babel) & Amharic (babel) \\
Spanish (hub) & Mandarin (hub) & Turkish (hub) \\
Turkish (hub) & English (libri) & Mandarin (hub) \\
Zulu (babel) & Swahili (babel) & Amharic (babel) \\
\bottomrule
\end{tabular}
\label{table:relation}
\end{center}
\end{table}

To understand how our corpus embeddings contribute to our approach, we rank those embeddings and show the top-2 similar corpora for each corpus in Table.\ref{table:relation}. We note that the target corpus itself is removed from the rankings because it is the most related corpus to itself. The results of the top half show very clearly that our embeddings can capture the language level information: For most English and Mandarin corpora, the most related corpus is another English or Mandarin corpus. Additionally, the bottom half of the table indicates that our embeddings are able to capture domain level information as well. For instance, the top 2 related corpus for Amharic is Bengali and Swahili. According to Table.\ref{tab:corpus}, those three corpora belong to the \textit{telephone} domain. In addition, Dutch is a \textit{read} corpus, its top 2 related corpora are also from the same domain. This also explains why the 1st related corpus of Mandarin (hk) is Bengali: because both of them are from the same \textit{telephone} domain. 

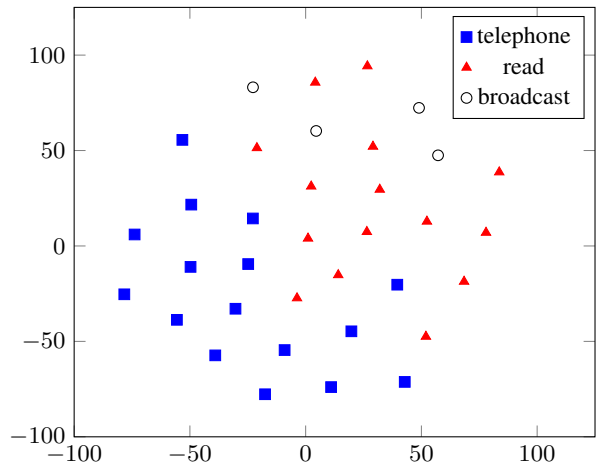
\begin{figure}
\centering
\begin{tikzpicture}
	\begin{axis}[%
	scatter/classes={%
		telephone={mark=square*,blue},%
		record={mark=triangle*,red},%
		broadcast={mark=o,draw=black}},	xmin=-100,xmax=125,ymin=-100,ymax=125]
	\addplot[scatter,only marks,%
		scatter src=explicit symbolic]%
	table[meta=label] {
	
x                  y                  label
-30.31571388244629 -32.897972106933594 telephone
2.3990633487701416 31.22985076904297 record
11.002802848815918 -73.944580078125 telephone
51.9562873840332 -47.479148864746094 record
-24.818572998046875 -9.482719421386719 telephone
-9.081963539123535 -54.546974182128906 telephone
-21.057504653930664 51.35031509399414 record
-55.622982025146484 -38.70508575439453 telephone
4.542168617248535 60.26102828979492 broadcast
77.91976165771484 7.000478744506836 record
68.43834686279297 -18.6563777923584 record
-49.72146224975586 -10.975892066955566 telephone
52.388553619384766 12.837336540222168 record
1.0258090496063232 3.9491748809814453 record
-3.7222859859466553 -27.329965591430664 record
39.575313568115234 -20.297710418701172 telephone
-22.801538467407227 14.41296100616455 telephone
-78.33529663085938 -25.326885223388672 telephone
14.079936027526855 -15.207659721374512 record
19.753564834594727 -44.717838287353516 telephone
42.81153869628906 -71.23609924316406 telephone
48.969696044921875 72.3545913696289 broadcast
83.62925720214844 38.655426025390625 record
-49.41233825683594 21.66193199157715 telephone
26.44234848022461 7.403919219970703 record
-73.8827896118164 6.017458438873291 telephone
-17.529891967773438 -77.6971664428711 telephone
-39.04698181152344 -57.32247543334961 telephone
29.086034774780273 52.09471893310547 record
32.02229309082031 29.5660457611084 record
57.19882583618164 47.49128723144531 broadcast
26.665611267089844 94.22811126708984 record
4.1577959060668945 85.670166015625 record
-22.751380920410156 83.13924407958984 broadcast
-53.254608154296875 55.57710647583008 telephone
		};
		\legend{telephone, read, broadcast}
	\end{axis}
\end{tikzpicture}
	\caption{Domain plot of 36 corpora, the corpus embeddings are reduced to 2 dimensions by t-SNE}
		\label{fig:domain}
\end{figure}

To further investigate the domain information contained in the corpus embeddings, we train the corpus embeddings with an even larger corpora collection (36 corpora) and plot all of them in Figure.\ref{fig:domain}. To create the plot, the dimension of each corpus embedding is reduced to 2 with t-SNE \cite{maaten2008visualizing}. The figure demonstrates clearly that our corpus embeddings are capable of capturing the domain information: all corpora with the same domain are clustered together. This result also means that our approach improves the model by sampling more frequently from the corpora of the same speech domain.

\section{Conclusion}
\label{sec:conclusion}
In this work, we propose an approach to compute corpus-level embeddings. We also introduce Corpus Relatedness Sampling approach to train multilingual speech recognition models based on those corpus embeddings. Our experiment shows that our approach outperforms the fine-tuned multilingual models in all 16 test corpora by 1.6 phone error rate on average. Additionally, we demonstrate that our corpus embeddings can capture both language and domain information of each corpus.

\section{Acknowledgements}
This project was sponsored by the Defense Advanced Research Projects Agency (DARPA) Information Innovation Office (I2O), program: Low Resource Languages for Emergent Incidents (LORELEI), issued by DARPA/I2O under Contract No. HR0011-15-C-0114. 

\bibliographystyle{IEEEtran}
\bibliography{mybib}

\end{document}